\pdfoutput=1

\documentclass[11pt]{article}

\usepackage[preprint]{acl}

\usepackage{times}
\usepackage{latexsym}
\usepackage[T1]{fontenc}

\usepackage[utf8]{inputenc}

\usepackage{microtype}

\usepackage{inconsolata}

\usepackage{enumitem}
\usepackage{colortbl} 
\usepackage{xcolor}   
\usepackage{algorithm}
\usepackage{algorithmic}

\usepackage{amsmath,amsthm,amsfonts,amssymb}
\usepackage{bm}
\usepackage{graphicx}
\usepackage{sidecap}
\usepackage{mathrsfs}
\usepackage{multirow, multicol}
\usepackage{caption}
\usepackage{wrapfig}
\usepackage{booktabs} 
\usepackage{xcolor}
\usepackage{subcaption}

\def\eqref#1{equation~\ref{#1}}

\def\1{\bm{1}}

\def\va{{\bm{a}}}

\def\vc{{\bm{c}}}

\def\vh{{\bm{h}}}

\def\vs{{\bm{s}}}

\def\vv{{\bm{v}}}

\def\vx{{\bm{x}}}
\def\vy{{\bm{y}}}

\def\mA{{\bm{A}}}

\def\mH{{\bm{H}}}

\def\mX{{\bm{X}}}
\def\mY{{\bm{Y}}}

\DeclareMathAlphabet{\mathsfit}{\encodingdefault}{\sfdefault}{m}{sl}
\SetMathAlphabet{\mathsfit}{bold}{\encodingdefault}{\sfdefault}{bx}{n}

\def\gC{{\mathcal{C}}}

\def\gO{{\mathcal{O}}}

\def\gS{{\mathcal{S}}}

\newcommand{\E}{\mathbb{E}}

\title{Chunk, Align, Select: A Simple Long-sequence Processing Method for Transformers}

\author{\textbf{Jiawen Xie$^{12}$\thanks{Work done during an internship at Tencent AI Lab.}, Pengyu Cheng$^{1}$\thanks{Corresponding author.}, Xiao Liang$^{13}$, Yong Dai$^{1}$, Nan Du$^{1}$} \\
$^{1}$Tencent AI Lab \\
$^{2}$Shanghai Jiao Tong University \\
$^{3}$Tsinghua University}

\begin{document}
\maketitle
\begin{abstract}
Although dominant in natural language processing, transformer-based models still struggle with long-sequence processing, due to the computational costs of their self-attention operations, which increase exponentially as the length of the input sequence grows. To address this challenge, we propose a \textbf{Sim}ple framework to enhance the long-content processing of off-the-shelf pre-trained transformers via three steps: \textbf{C}hunk, \textbf{A}lign, and \textbf{S}elect (SimCAS). More specifically, we first divide each long-sequence input into a batch of chunks, then align the inter-chunk information during the encoding steps, and finally, select the most representative hidden states from the encoder for the decoding process. 
With our SimCAS, the computation and memory costs can be reduced to linear complexity. In experiments, we demonstrate the effectiveness of the proposed method on various real-world long-text summarization and reading comprehension tasks, in which SimCAS significantly outperforms prior long-sequence processing baselines. The code is at \url{https://github.com/xjw-nlp/SimCAS}. 
\end{abstract}

\section{Introduction} \label{sec:introduction}
Transformers \citep{NIPS2017_3f5ee243} have become a fundamental model architecture for sequential data modeling \citep{sun2019videobert,dosovitskiy2021an,10.1162/coli_a_00462}, especially in  Natural Language Processing (NLP) \citep{devlin2018bert,NEURIPS2020c1457c0d6}, where texts are regarded as sequences of tokens.  Built with transformer blocks, pre-trained language models (PLMs) have recently shown astonishing empirical performance in various NLP tasks such as question answering \cite{yang-etal-2019-end-end}, controllable generation \citep{byrne-etal-2021-tickettalk,cheng2022replacing}, summarization \citep{rush-etal-2015-neural,nallapati-etal-2016-abstractive} and logic reasoning~\citep{wei2022chain,cheng2024spag}. However, one fatal weakness, that has hindered transformer-based models from being applied in broader application scenarios, is the quadratically raised computational consumption of self-attention operations when the input length increases. Hence, vanilla transformers have continuously been challenged by long-context tasks, such as machine reading comprehension \citep{10.1162/tacl_a_00276,gong-etal-2020-recurrent,pang-etal-2022-quality} and long-text summarization~\citep{huang-etal-2021-efficient,10.1145/3529754}. 

To enhance transformers with more efficient long-sequence processing, 
prior works focus on two perspectives, efficient attention operations~\citep{beltagy2020longformer,zaheer2020big,choromanski2020rethinking}  and sub-sequence processing~\citep{liu-etal-2022-leveraging-locality}. 
Efficient attention targets on reducing the memory and calculation cost of self-attention operations while preserving transformers' empirical performance on downstream tasks. Unfortunately, most efficient attention methods require customized self-attention implementations, which demand from-scratch training instead of being directly plugged into existing pre-trained models. 
Moreover, some empirical studies have demonstrated that efficient-attention methods inevitably sacrifice the short-sequence processing performance compared with full-attention models~\citep{phang2022investigating}.

The alternative solution to long-sequence processing is to decompose the long-sequence input into multiple sub-sequences and then feed-forward each separately, known as sub-sequence processing~\citep{moro-etal-2022-discriminative}. Although full-attention blocks are sufficiently utilized in each sub-sequence, these methods have been challenged to capture the semantic information across different sub-contexts. 
To solve this, some works assign the same fragment to different chunks~\citep{10.1162/tacl_a_00547}, which however significantly increases the computational cost of each chunk and runs counter to the original efficiency goal. 

To gather the advantages of the methods above, we introduce a \textbf{Sim}ple yet effective learning framework with three typical operations: \textbf{C}hunk, \textbf{A}lign, and \textbf{S}elect (\text{SimCAS}). In detail, SimCAS first chunks the input sequence into a batch of sub-sequences then feeds each subsequence forward the elaborately designed encoding blocks with an inter-chunk alignment mechanism. Finally, the most semantically representative hidden representations are selected via a tailored selection policy to compress the overall sequence length. To align the semantic information across sub-sequence chunks, we introduce a sequential batch alignment operation to calibrate the start and end token embeddings of each sub-sequence in the encoder layers. To learn the selector policy, inspired by the recent success of reinforcement learning in NLP~\citep{ouyang2022training}, we adopt 
the Proximal Policy Optimization (PPO) \citep{DBLP:journals/corr/SchulmanWDRK17} algorithm with the decoder treated as the environment. Moreover, we use the attention scores and output likelihood to calculate effective rewards for the selector's policy optimization.
To verify the effectiveness of SimCA,
we conducted comprehensive experiments on seven long-context datasets from the domains of document-level summarization, multi-document summarization, and reading comprehension. 
The empirical results show that SimCAS outperforms other long-sequence processing baselines with high-level scalability. The main contributions of this paper are:
\begin{itemize}[leftmargin=0.4cm]
\item  We propose a simple yet effective long-sequence processing framework, which noticeably extends the application range of existing full-attention PLMs. Unlike prior works compromising the performance on short sequences, our method maintains satisfying performances processing either short or long contexts.
\item We discover that transformers can be conceptualized as simulation environments for policy learning. We leverage transformers' attention scores and output likelihoods to optimize the selector policy to compress the long-sequence information. The optimized selector policy meanwhile facilitates the transformer to concentrate more on hidden states with high semantic importance. 
\item We conduct comprehensive experiments illustrating that SimCAS consistently surpasses previous baselines. Furthermore, we provide model scaling, efficiency comparisons, and ablation studies to substantiate the superior performance of our proposed method.
\end{itemize}

\section{Background} \label{sec:preliminary}
\paragraph{Language Modeling}
The training objective for sequence generation consists of a sequence of token decisions made in an auto-regressive manner. This is formulated as a product of decision probabilities corresponding to specific tokens. Given an input sequence $\mX = (\vx_{1}, \vx_{2},\cdots, \vx_N)$ and its corresponding output $\mY = (\vy_{1}, \vy_{2},\cdots, \vy_M)$, we model the following conditional probability:
\begin{equation} \label{eq_2}  
p_{\phi}(\mY |\mX) = \prod_{m = 1}^{M} p_\phi(\vy_{m} | \mY_{<m}, \mX),
\end{equation}
where $\mY_{<m} = (\vy_1, \vy_2, \dots, \vy_{m-1})$, and  $\phi$ represents the model parameters.

\paragraph{Proximal Policy Optimization}
In the domain of reinforcement learning (RL), Proximal Policy Optimization (PPO)~\citep{DBLP:journals/corr/SchulmanWDRK17} is a widely used policy gradient method~\citep{kakade2001natural} for its remarkable performance and efficiency in solving complex control and decision-making tasks~\citep{Vinyals2019GrandmasterLI,akkaya2019solving,cheng2024adversarial}.
The vanilla policy gradient estimator has the form:
$    \nabla_{\theta} \E_{\pi_\theta(\va_t|\vs_t)} [A^\pi_t(\va_t, \vs_t)] \approx  \hat{\E}_t [\nabla_\theta \log \pi_\theta(\va_t | \vs_t) \hat{A}_t ]$, where $\vs_t\in \gS$  is the state at $t$-step, $\pi_\theta(\va_t | \vs_t)$ is a stochastic policy acting $\va_t$ at $\vs_t$, $\hat{A}_t$ is the estimated value of the advantage function $A_t^\pi(\va_t, \vs_t)$, and $\hat{\E}_t$ denotes the empirical average over a sample batch. The PPO algorithm improves the training stability of the policy gradient, by optimizing the following objective:
\begin{equation} %
L = \hat{\E}_t [\min(r_t(\theta)\hat{A}_t, \text{clip}(r_t(\theta), 1- \varepsilon, 1 + \varepsilon) \hat{A}_t)],
\end{equation}
where $r_t(\theta) = \frac{\pi_\theta(\va_t| \vs_t)}{\pi_{\theta_\text{old}}(\va_t| \vs_t)}$ is the probability ratio between new and old policies, and $\varepsilon>0$ is a hyper-parameter for clipping.

\begin{figure*}[t]
    \centering
    \includegraphics[width=0.94\textwidth]{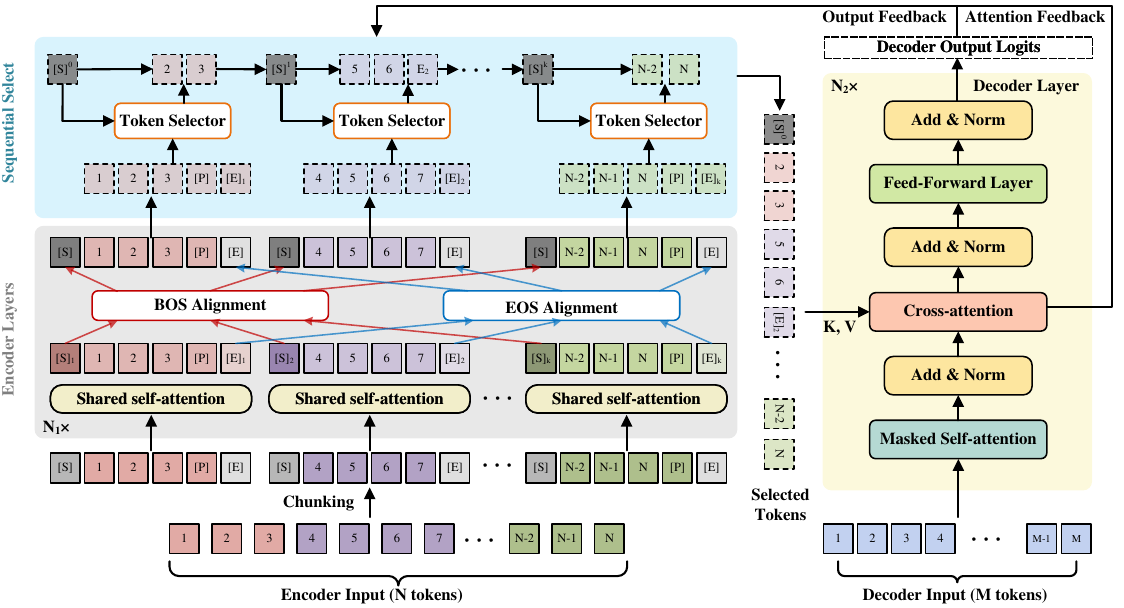}
    \caption{The SimCAS framework: The long inputs are first divided into a batch of chunks, each of which is filled with start token \texttt{[S]}, padding token \texttt{[P]} and end token \texttt{[E]}. Then the inter-chunk information can be transferred via the alignment of \texttt{[S]} and \texttt{[E]} representations after each encoder layer. Next, hidden states are selected for decoding steps. The decoder output logits and attention scores are utilized as rewards for updating the token selector.}
    \label{fig:pipeline}
\end{figure*}

\section{Methodology}
Given a long-input text $\mX=(\vx_1, \vx_2, \dots, \vx_N)$ with a fairly large input length $N$, we aim to design a model $p_\phi(\mY | \mX)$ to predict a corresponding label sequence $\mY = (\vy_1, \vy_2, \dots, \vy_M)$, where $\mY$ can either be  classification labels or output sequence tokens. The major difficulty of the task is that the input length $N$ is so large that the original self-attention operations become infeasible with the quadratic complexity $\gO(N^2)$.

To address the challenge, we propose a novel method that intuitively splits the long inputs into chunks with feasible lengths, then selects the most representative tokens for decoding steps. To guarantee inter-chunk semantic information extracted during encoding, we design an aligning scheme in the encoder blocks. In the following, we will describe the chunking scheme, the aligning strategy, and the selector design in detail.

\subsection{Chunking Scheme}

Assume the maximum input sequence length of a pre-trained transformer model is $S$. We first split the long input sequence into $B = \lceil \frac{N}{S} \rceil$ chunks with each chunk length equal to $S$. To simplify the expression, we assume that a sentence is in only one chunk. In the practical experiment, we use the sentence-level segmentation \citep{Moro_Ragazzi_2022} to acquire a sequence of sentences, and then employ the greedy method to allocate sequentially these sentences to the chunks. If there is not enough space in the chunk for the current sentence, we will use the padding tokens to fill the chunk. Mathematically, these divided segments can be represented as:
%
%
\begin{equation}
\big\{(\vc^k_{1}, \vc^k_{2}, \dots, \vc^k_{S})\big\}_{k=1}^{B},
\end{equation} 
where $\vc^k_i = \vx_{(k-1)S+i}$, and $\lceil \cdot \rceil$ is the ceiling function. Since the chunk might not have $S$ tokens from $\mX$, we append special padding tokens at the end (see Figure~\ref{fig:pipeline}). After chunking, we add the start token (\texttt{[S]}) and end token (\texttt{[E]}) to each chunk, and treat the chunks as a normal transformer input batch $\gC$ with batch size $B$:
\begin{equation}\label{eq:chunks-with-special-tokens}
    \gC = \big\{(\texttt{[S]}, \vc^k_{1}, \vc^k_{2} \dots, \vc^k_{S}, \texttt{[E]})\big\}_{k=1}^{B}.
\end{equation}

\subsection{Sequential Batch Alignment}

After chunking the input text into a standard token batch, we can encode it with the transformer encoder layers. As in Figure~\ref{fig:pipeline}, we assume the encoder has $N_1$ layers. Denote the hidden representations of $k$-th chunk in $\gC$ (in \eqref{eq:chunks-with-special-tokens}) at $l$-th encoder layer as $\mH^{k,l} = (\vh^{k,l}_0, \vh^{k,l}_1, \dots, \vh^{k,l}_S, \vh^{k,l}_{S+1})$, where $\vh^{k,l}_0$ and $\vh^{k,l}_{S+1}$ are hidden representation of \texttt{[S]} and \texttt{[E]} tokens respectively, and $\vh^{k,l}_i$ is the embedding for $\vc^k_i$ with $1 \leq i \leq S$.

As mentioned in Section~\ref{sec:introduction}, chunking methods make it difficult to capture the inter-chunk semantic information. Inspired by recent work using special tokens for full sequence information \citep{mohtashami2023landmark}, we align the information of \texttt{[S]} and \texttt{[E]} tokens at each encoding block. More specifically, at $l$-th layer, our batch alignment average the hidden states of \texttt{[S]} and \texttt{[E]} of all chunks:
\begin{equation}
    \bar{\vh}^l_\text{BOS} = \frac{1}{B}\sum_{k=1}^B \vh^{k,l}_{0}, \ \ \bar{\vh}^l_\text{EOS} = \frac{1}{B} \sum_{k=1}^B \vh^{k,l}_{S+1}.
\end{equation}

Then we replace $\vh^{k,l}_{0}$ and $\vh^{k,l}_{S+1}$ with the aligned  $\bar{\vh}^l_\text{BOS}$ and $\bar{\vh}^l_\text{EOS}$ into the hidden states for the next-layer encoding block, as shown in Figure~\ref{fig:pipeline}.

\subsection{Token Selector}

After being encoded into the last hidden space with our sequential batch aligning scheme, the chunks should be reformed back to a sequence for the next decoding steps. Directly tiling the chunks' representations back into a sequence is still infeasible because the overlong sequence makes it difficult to fuse information at the decoding stage. Therefore, we propose a token selector to select the most representative hidden representations for decoding steps. Inspired by \citet{ramamurthy2023is}, we design the selector from the perspective of reinforcement learning (RL). 

\paragraph{Selection Module Design} Formally, the token selector takes the last hidden states $\mH^L = \{\vh^{k,L}_0, \vh^{k,L}_1, \dots, \vh^{k,L}_S, \vh^{k,L}_{S+1}\}_{k=1}^B$  and selection actions $\mA_t = (\va_1, \va_2, \dots, \va_{t-1})$ as inputs and predicts the next selection action $\va_{t}$, where each action $\va_t$ has two values \textit{"select"} and \textit{"skip"} for operating the $t$-th token $\vx_t$ in $\mX$.
We set the state of RL as $\vs_t = (\mH^L, \mA_t)$, then the selector is a policy $\pi_{\theta}(\va_t| \vs_t)$ to predict next action $\va_t$.

We implement the selector with the actor-critic style \citep{NIPS1999_6449f44a}. Both actor and critic are simple feed-forward networks, but the actor outputs a probability distribution over the action space and the critic outputs a single scalar value. At state $\vs_t$, to consider the sequential effect of previous action $\mA_t$, we first take the average of all selected hidden states as 
\begin{equation}
    \bar{\vh}_{t} = \frac{\sum_{k,i} \mathbb{I}_{\{\va_j = \textit{"select"}, j< t\}} \vh_i^{k,L}}{\sum_{k,i} \mathbb{I}_{\{\va_j = \textit{"select"}, j< t\}}},
\end{equation}\noindent
where $j=(k-1)S + i$ maps chunk indices back to the input sequence, and $\mathbb{I}_{\{\cdot\}}$ is the indicator function. Then we concatenate the current selector state $\bar{\vh}_t$ and token information $\vh_i^{k,L}$, $t=(k-1)S+i$, to predict next action $\va_t$ via the actor:
\begin{equation}%
    \pi_{\theta}(\va_t|\vs_t) = \text{actor}(\bar{\vh}_t, \vh_i^{k,L}).
\end{equation}\noindent

\paragraph{Reward Design}
To train the selector within an RL scheme, we treat the transformer as an environment and design action rewards. Inspired by \citet{NEURIPS2021_e4d2b6e6}, we can directly utilize the language modeling likelihood as the generation quality reward for selection actions:
\begin{equation}  
    R_\text{LM} = \xi \exp \{\frac{1}{M} \log p_{\phi}(\mY |\mX)\},
\end{equation}\noindent
where $\xi$ is a coefficient that magnifies the value of the reward for easier optimization of the selector.
However, $R_\text{LM}$ is only a scalar value, which cannot provide fine-grained guidance to the selector. Therefore, we use the input-output cross-attention scores to calibrate $R_\text{LM}$. More specifically, 
%
%
we denote the cross-attention matrix of the $q$-th attention head in the $l$-th layer of the decoder as $\mA^{l}_{q} \in \mathbb{R}^{M \times N}$, and the overall cross-attention
\begin{equation} 
\bar{\mA} = \frac{1}{N_2 \cdot Q}\sum_{l = 1}^{N_2}\sum_{q = 1}^{Q}\mA^{l}_{q},
\end{equation}
where $N_2$ is the number of decoder layers, and $Q$ is the number of cross-attention heads. With the overall feedback $R_\text{LM}$ and cross-attention  $\bar{\mA} = (\bar{a}_{ij})_{M \times N}$, we adjust the reward $R_{j}^{+}$ to each selected tokens. For \textit{"skip"} action, we intend to limit the selected sequence length. Assume the number of overall input tokens and selected token representations is ${L}_\text{all}$ and ${L}_\text{select}$ respectively. We set a hyper-parameter ${L}_\text{hyper}$ to control  the size of ${L}_\text{select}$ with the skipping reward $R^{-}$:
\begin{equation}  
\begin{aligned}\label{eq:select_skip_reward} 
&R_{j}^{+} = \frac{\bar{a}_{j}}{1 - \bar{a}_{0}} R_\text{LM}, \bar{a}_{j} = \frac{1}{M} \sum_{i = 1}^{M} \bar{a}_{ij},\\
&R^{-} = 
\begin{cases}
\frac{R_\text{LM}}{L_\text{all}} & \text{if ${L}_\text{select} < {L}_\text{hyper}$} \\
\frac{R_\text{LM}}{L_\text{select}} & \text{otherwise.}
\end{cases} \\
\end{aligned}
\end{equation}

With the selector and rewards designed above, we can optimize the selector with the PPO algorithm described in Section~\ref{sec:preliminary}. Note that in our setups, the environment (the transformer) changes during the training steps. Therefore, we alternatively update the selector and transformer: in each interaction,  we first fix the transformer and use the reward $R_\text{LM}$ and cross-attention scores to update the selector, then fix the selector and update the transformer with language modeling loss. In addition, the selector is similar to the chunk-wise RNN, so the time overhead of the selection process is low.

\section{Related Work}
\paragraph{Efficient Transformers}
The attention mechanism in transformers requires quadratically increased computational complexity with respect to the input sequence length, which limits the application scenarios, especially for long-text processing.  
To address this issue, various previous works have been proposed for designing more efficient attention operations \citep{tay2023scaling,10.1145/3586074}. Longformer~\citep{beltagy2020longformer}, BIGBIRD~\citep{zaheer2020big}, GMAT \citep{gupta2020gmat}, and ETC~\citep{ainslie-etal-2020-etc} reduce the memory consumption of dense attentions by elaborate combinations of global attention and local attention mechanisms. LongT5 \citep{guo-etal-2022-longt5} is based on the original T5 with global-local attention sparsity patterns to handle long input and has an additional pre-training phase on C4 dataset. CoLT5 \citep{ainslie-etal-2023-colt5} is an improved version of LongT5, with the same parameter size and pre-training dataset as LongT5, and it optimizes the computation of the attention module and the feed-forward module and proposes a new pre-training objective.
Additionally, Reformer~\citep{Kitaev2020Reformer:} leverages a locality-sensitive hashing to the attention mechanism, changing its complexity from $\mathcal{O}(n^{2})$ to $\mathcal{O}(n\log n)$, where $n$ is the input text sequence length. Routing Transformer~\citep{10.1162/tacl_a_00353} applies a sparse routing module based on online k-means to self-attention while reducing the overall complexity of attention. Approximation-based methods, such as Performers~\citep{choromanski2020rethinking} and RFA~\citep{peng2021random}, use linear space and time complexity to estimate the attention matrix based on random features. Luna~\citep{ma2021luna} attends only to a fixed number of hidden vectors. Linformer \citep{wang2020linformer} calculates self-attention by a low-rank matrix. However, the vast majority of these methods are difficult to apply to existing PLMs. Moreover, \citet{xiong-etal-2022-simple} proposes that many efficient-attention transformers do not even perform as well as simple local-attention models on downstream language tasks.

\paragraph{Chunking Methods for Long Sequence}
Another solution for long-sequence processing is to perform sequence chunking and then process them respectively \citep{zhong-etal-2022-training,wang2023augmenting}.
Among chunking methods, SLED~\citep{10.1162/tacl_a_00547} splits the long sequence into overlapping chunks and processes each chunk with the encoder, then fuses cross-chunk information with the decoder. PageSum~\citep{liu-etal-2022-leveraging-locality} separates the long sequence into non-overlapping chunks and effectively tackles them by the principle of locality~\citep{10.1145/1070838.1070856}. Unlimiformer~\citep{bertsch2023unlimiformer} encodes long inputs in chunks and utilizes only the top-k input tokens for every attention head.

\paragraph{Length Extrapolation} Length extrapolation in transformers refers to their ability to handle input sequences of varying lengths during both training and inference \citep{press2022train,sun-etal-2023-length}. Transformers use a self-attention mechanism to capture dependencies across positions in a sequence, allowing them to generalize well to sequences of different lengths. This flexibility is essential for tasks like NLP and time series analysis, where input lengths can vary.

\paragraph{Sequence Length Reduction} Reducing the length of hidden states \citep{guan-etal-2022-transkimmer,kim2022learned} is the method of model compression from the width perspective, which is promising since some studies showed that there is redundant encoded information in token representations \citep{ethayarajh2019contextual,klafka-ettinger-2020-spying}. Among the redundancy, some tokens carry more task-specific information than others, suggesting that these tokens are more salient and imperative to be selected to feed into subsequent operations. Compared with model compression via layer-wise pruning, token-level pruning does not come at the expense of model performance in complex reasoning \citep{sanh2019distilbert,sun2019patient}.

\begin{table*}[tbp] \small
    \centering
    \setlength{\tabcolsep}{3.2mm}
    \begin{tabular}{ l l c c c c c c c c}
        \toprule
         \multirow{2}*{\textbf{Base Model}} & \multirow{2}*{\textbf{Method}} & \multicolumn{4}{c}{\textbf{arXiv}}  & \multicolumn{4}{c}{\textbf{PubMed}} \\
         \cmidrule(lr){3-6} \cmidrule(lr){7-10} 
         ~ & ~ & \textbf{R-1} & \textbf{R-2} & \textbf{R-L} & \textbf{BS} & \textbf{R-1} & \textbf{R-2} & \textbf{R-L} & \textbf{BS} \\
        \midrule
        LED$_\text{large}$ & Standard & 46.63 & 19.62 & 41.83 & - & - & - & - & - \\
        LED$_\text{large}$ & PRIMERA & 47.60 & \textbf{20.80} & \underline{42.60} & - & - & - & - & - \\
        PEGASUS$_\text{large}$ & Standard & 44.21 & 16.95 & 38.83 & - & 45.97 & 20.15 & 41.34 & - \\
        PEGASUS$_\text{large}$ & BIGBIRD & 46.63 & 19.02 & 41.77 & - & 46.32 & 20.65 & 42.33 & - \\
        BART$_\text{large}$ & HEPOS & \underline{47.87} & \underline{20.00} & 41.50 & - & 47.93 & 20.74 & 42.58 & - \\
        \midrule
        BART$_\text{base}$ & Standard & 40.36 & 13.78 & 36.11 & 59.44 & 40.36 & 13.29 & 35.02 & 61.77 \\
        BART$_\text{large}$ & Standard & 42.97 & 15.54 & 37.02 & 61.18 & 42.87 & 15.44 & 36.93 & 63.08 \\
        BART$_\text{base}$ & SimCAS & 47.22 & 19.35 & 42.25 & \underline{63.51} & \underline{48.17} & \underline{21.11} & \underline{43.90} & \underline{66.33} \\
        BART$_\text{large}$ & SimCAS & \textbf{48.14} & 19.77 & \textbf{42.93} & \textbf{63.78} & \textbf{48.65} & \textbf{21.40} & \textbf{44.14} & \textbf{66.52} \\
        \bottomrule
        \end{tabular}
        \caption{Average results on arXiv and PubMed test sets. R-1/2/L is the ROUGE-1/2/L F1 score. BS refers to the neural model-based metrics BERTScore. Bold (underline) are used to denote the best (second-best) metric.}
        \label{tab:long}
\end{table*}

\vspace{-1mm}
\section{Experiments}
\vspace{-1mm}
To evaluate the performance of SimCAS, we conduct experiments on the long-text abstractive summarization and machine reading comprehension tasks. In the following, we introduce the information about the datasets, baselines, model implementations, and evaluation results of our experiments.

\subsection{Datasets}
\vspace{-1mm}
We conduct experiments on two types of NLP tasks: long-text summarization and machine reading comprehension. For long-text summarization, we use four single-document summarization datasets: arXiv, PubMed~\citep{cohan-etal-2018-discourse}, GovReport \citep{huang-etal-2021-efficient}, SummScreen \citep{chen-etal-2022-summscreen}, and two multi-document summarization datasets: Multi-News~\citep{fabbri2019multi}, WCEP~\citep{gholipour-ghalandari-etal-2020-large}.
For the reading comprehension task, we test on the NarrativeQA~\citep{10.1162/tacl_a_00023} dataset. We list the introduction about these datasets below. More dataset details can be found in Appendix \S~\ref{appx:statistics}. 

{\bf arXiv \& PubMed}\footnote{\href{https://github.com/armancohan/long-summarization}{https://github.com/armancohan/long-summarization}} are two long-document summarization datasets in the scientific research domain. Each document is a scientific paper whose summary is the corresponding abstract.

{\bf GovReport}\footnote{\href{https://github.com/luyang-huang96/LongDocSum}{https://github.com/luyang-huang96/LongDocSum}} is a long-document summarization dataset based on reports published by the U.S. Government Accountability Office and Congressional Research Service.

{\bf SummScreen}\footnote{\href{https://github.com/mingdachen/SummScreen}{https://github.com/mingdachen/SummScreen}}, which includes TV series transcripts, often indirectly presents plot details through character dialogues scattered throughout the transcript. These details need to be consolidated to create concise plot descriptions.

{\bf Multi-News}\footnote{\href{https://github.com/Alex-Fabbri/Multi-News}{https://github.com/Alex-Fabbri/Multi-News}} is a large-scale multi-document summarization dataset. It consists of news articles and human-written summaries of these articles. Each summary is professionally written by editors and with links to the original articles cited.

{\bf WCEP}\footnote{\href{https://github.com/allenai/PRIMER}{https://github.com/allenai/PRIMER}} is a dataset for multi-document summarization (MDS). It contains short, human-written summaries about news events, obtained from the Wikipedia Current Events Portal (WCEP). 

{\bf NarrativeQA}\footnote{\href{https://github.com/deepmind/narrativeqa}{https://github.com/deepmind/narrativeqa}} is a reading comprehension dataset over entire books from Project Gutenberg and movie scripts from different websites.


\subsection{Baselines}
There are several baselines for comparison: among them, those based on the full-attention mechanism are HiMAP \citep{fabbri-etal-2019-multi}, BERTREG \citep{gholipour-ghalandari-etal-2020-large},  {Submodular+Abs} \citep{gholipour-ghalandari-etal-2020-large}, BART~\citep{lewis-etal-2020-bart}, PEGASUS~\citep{pmlr-v119-zhang20ae}, DynE \citep{hokamp2020dyne}, GraphSum \citep{li-etal-2020-leveraging-graph}, BART-Long-Graph \citep{pasunuru-etal-2021-efficiently}, SLED~\citep{10.1162/tacl_a_00547}, Memorizing Transformers \citep{wu2022memorizing}, Unlimiformer \citep{bertsch2023unlimiformer}.

Moreover, several baselines introduce the tailored sparse attention to address longer sequences such as LED~(Longformer Encoder-Decoder) \citep{beltagy2020longformer},  BIGBIRD~\citep{zaheer2020big}, {PRIMERA} \citep{xiao-etal-2022-primera}, HEPOS \citep{huang-etal-2021-efficient} and LED+RELAX~\citep{parnell2022multidocument}. More details of baselines can be found in Appendix \S \ref{appx:baseline}.

\begin{table}[tbp]
    \centering
    \setlength{\tabcolsep}{3.0mm}
    \begin{tabular}{ c c c c c}
         ~ & \textbf{R-1} & \textbf{R-2} & \textbf{R-L} & \textbf{BS} \\
        \textbf{R-1} & \cellcolor{purple!5}1.000 & \cellcolor{purple!20}0.872 & \cellcolor{purple!10}0.913 & \cellcolor{purple!30}0.810  \\
        \textbf{R-2} & \cellcolor{purple!20}0.872 & \cellcolor{purple!5}1.000 & \cellcolor{purple!15}0.898 & \cellcolor{purple!35}0.794 \\
        \textbf{R-L} & \cellcolor{purple!10}0.913 & \cellcolor{purple!15}0.898 & \cellcolor{purple!5}1.000 & \cellcolor{purple!25}0.825 \\
        \textbf{BS} & \cellcolor{purple!30}0.810 & \cellcolor{purple!35}0.794 & \cellcolor{purple!25}0.825 & \cellcolor{purple!5}1.000 \\
        \end{tabular}
        \caption{\textbf{Pearson correlation coefficient} between the four automatic evaluation metrics (R-1, R-2, R-L, BS) used for a base BART with beam search on \textbf{GovReport}.}
        \label{tab:pearson}
\end{table}
\subsection{Implementation Details}

\begin{table*}[tbp] \small
    \centering
    \setlength{\tabcolsep}{3.2mm}
    \begin{tabular}{ l l c c c c c c c c}
        \toprule
         \multirow{2}*{\textbf{Base Model}} & \multirow{2}*{\textbf{Method}} & \multicolumn{4}{c}{\textbf{GovReport}}  & \multicolumn{4}{c}{\textbf{SummScreen}} \\
         \cmidrule(lr){3-6} \cmidrule(lr){7-10} 
         ~ & ~ & \textbf{R-1} & \textbf{R-2} & \textbf{R-L} & \textbf{BS} & \textbf{R-1} & \textbf{R-2} & \textbf{R-L} & \textbf{BS} \\
        \midrule
        BART$_\text{base}$ & SLED & 54.70 & 24.40 & 25.40 & - & 32.70 & 7.90 & 19.10 & - \\
        BART$_\text{large}$ & SLED & 57.50 & 26.30 & 27.40 & - & 35.20 & 8.70 & \underline{19.40} & - \\
        LED$_\text{large}$ & PRIMERA & 55.10 & 23.90 & 25.90 & 67.00 & 32.30 & 7.10 & 18.30 & 57.10 \\
        BART$_\text{base}$ & Memorizing & 55.20 & 25.10 & 26.40 & 67.50 & 32.70 & 7.40 & 19.20 & 57.40 \\
        BART$_\text{base}$ & Unlimiformer & 56.60 & \underline{26.30} & 27.60 & \underline{68.20} & 34.70 & 8.50 & \textbf{19.90} & 58.50 \\
        PRIMERA & Unlimiformer & 57.40 & 26.20 & \textbf{28.00} & 68.10 & 33.30 & 7.60 & 18.90 & 57.70 \\
        \midrule
        BART$_\text{base}$ & Standard & 51.72 & 19.37 & 23.11 & 64.12 & 29.73 & 5.23 & 15.65 & 54.30 \\
        BART$_\text{large}$ & Standard & 52.94 & 19.78 & 23.71 & 64.44 & 29.89 & 5.32 & 15.71 & 54.43 \\
        BART$_\text{base}$ & SimCAS & \underline{59.30} & 25.95 & 27.07 & 68.17 & \underline{43.45} & \underline{12.74} & 18.38 & \underline{62.46}\\
        BART$_\text{large}$ & SimCAS & \textbf{60.29} & \textbf{26.68} & \underline{27.97} & \textbf{68.64} & \textbf{44.15} & \textbf{13.42} & 18.50 & \textbf{62.82} \\
        \bottomrule
        \end{tabular}
        \caption{Average results on GovReport and SummScreen test sets. R-1/2/L is the ROUGE-1/2/L F1 score. BS refers to the neural model-based metrics BERTScore. The best and second-best results are bolded and underlined respectively.}
        \label{tab:long-1}
\end{table*}
\begin{table*}[tbp]
    \centering
    \setlength{\tabcolsep}{1.6mm}\small
    \begin{tabular}{ l c c c c}
        \toprule
         \textbf{System} & \textbf{R-1} & \textbf{R-2} & \textbf{R-L} & \textbf{BS} \\
        \midrule
        BART$_\text{large}$ & 42.04 & 14.88 & 23.34 & - \\
        HiMAP & 44.17 & 16.05 & 21.38 & - \\
        GraphSum & 45.87 & 17.56 & 23.39 & - \\
        BART-Long-Graph & 49.24 & 18.99 & 23.97 & - \\
        LED+RELAX & 47.23 & 18.86 & 25.03 & - \\
        PRIMERA & \textbf{49.94} & \textbf{21.05} & \underline{25.85} & - \\
        \midrule
        BART$_\text{base}$ & 40.54 & 12.18 & 22.39 & 58.75 \\
        BART$_\text{large}$ & 42.16 & 14.69 & 23.51 & 60.70 \\
        BART$_\text{base}$+SimCAS & 48.88 & 20.01 & 25.32 & \underline{65.05} \\
        BART$_\text{large}$+SimCAS & \underline{49.40} & \underline{20.47} & \textbf{25.96} & \textbf{65.40} \\
        \bottomrule
        \end{tabular}
        \quad
        \hspace{1.5em}
    \begin{tabular}{ l c c c c}
        \toprule
         \textbf{System} & \textbf{R-1} & \textbf{R-2} & \textbf{R-L} & \textbf{BS} \\
        \midrule
        BERTREG & 35.00 & 13.50 & 25.50 & - \\
        SUBMODULAR+ABS & 34.40 & 13.10 & 25.00 & - \\
        DynE & 35.40 & 15.10 & 25.60 & - \\
        LED & 39.79 & 18.94 & 32.10 & - \\
        LED+RELAX & 41.11 & 19.46 & 33.13 & - \\
        PRIMERA & \underline{46.08} & \textbf{25.21} & \underline{37.86} & - \\
        \midrule
        BART$_\text{base}$ & 36.12 & 13.77 & 29.98 & 60.84 \\
        BART$_\text{large}$ & 37.66 & 15.98 & 31.01 & 62.32 \\
        BART$_\text{base}$+SimCAS & 45.68 & 22.80 & 37.71 & \underline{70.59} \\
        BART$_\text{large}$+SimCAS & \textbf{46.29} & \underline{24.45} & \textbf{38.61} & \textbf{71.38} \\
        \bottomrule
        \end{tabular}
        \caption{Average results on Multi-News (left) and WCEP (right) test sets. R-1/2/L is the ROUGE-1/2/L F1 score. BS refers to the model-based metrics BERTScore. Bold (underline) are used to denote the best (second-best) metric.}
        \label{tab:multi-document}
\end{table*}

\begin{table*}[tbp]\small
    \centering
    \setlength{\tabcolsep}{2.0mm}
    \begin{tabular}{ l c c c c c c c}
        \toprule
        Datasets & w/o Chunk & w/o Align & w/o Select & SimCAS & $\Delta_{\text{Chunk}}$ & $\Delta_{\text{Align}}$ & $\Delta_{\text{Select}}$ \\
        \midrule
        arXiv & 30.10 $\pm$ 0.73 & 36.12 $\pm$ 0.47 & 32.70 $\pm$ 0.62 & \textbf{36.46} $\pm$ \textbf{0.56} & $\uparrow\text{21.13\%}$ & $\uparrow\text{0.94\%}$ & $\uparrow\text{11.50\%}$\\
        PubMed & 29.63 $\pm$ 0.51 & 37.20 $\pm$ 0.64 & 34.32 $\pm$ 0.59 & \textbf{37.77} $\pm$ \textbf{0.51} & $\uparrow\text{27.47\%}$ & $\uparrow\text{1.53\%}$ & $\uparrow\text{10.05\%}$ \\
        GovReport & 31.23 $\pm$ 0.50 & 37.12 $\pm$ 0.66 & 32.91 $\pm$ 0.81 & \textbf{37.54} $\pm$ \textbf{0.48} & $\uparrow\text{20.20\%}$ & $\uparrow\text{1.13\%}$ & $\uparrow\text{14.07\%}$ \\
        SummScreen & 19.71 $\pm$ 0.49 & 20.69 $\pm$ 0.45 & 19.63 $\pm$ 0.62 & \textbf{25.15} $\pm$ \textbf{0.42} & $\uparrow\text{27.60\%}$ & $\uparrow\text{2.56\%}$ & $\uparrow\text{08.10\%}$ \\
        Multi-News & 25.34 $\pm$ 0.38 & 31.45 $\pm$ 0.55 & 27.98 $\pm$ 0.35 & \textbf{31.36} $\pm$ \textbf{0.51} & $\uparrow\text{23.76\%}$ & $\downarrow\text{0.29\%}$ & $\uparrow\text{12.08\%}$ \\
        WCEP & 28.42 $\pm$ 0.72 & 35.01 $\pm$ 0.64 & 30.33 $\pm$ 0.76 & \textbf{35.31} $\pm$ \textbf{0.59} & $\uparrow\text{24.24\%}$ &$\uparrow\text{0.86\%}$ & $\uparrow\text{16.42\%}$ \\
        NarrativeQA & 21.70 $\pm$ 0.55 & 31.52 $\pm$ 0.46 & 23.20 $\pm$ 0.23 & \textbf{31.76} $\pm$ \textbf{0.44} & $\uparrow\text{46.36\%}$ & $\uparrow\text{0.76\%}$ & $\uparrow\text{36.90\%}$ \\
        \bottomrule
        \end{tabular}
        \caption{Ablation study results on the development sets of all datasets. Performance changes compared with the full model (BART$_\text{base}$+SimCAS) are reported. The metrics of summarization and reading comprehension datasets are the average of ROUGE-1/2/L and F1 respectively.}
        \label{tab:ablation}
\end{table*}
\begin{figure*}[tbp]
    \centering
    \includegraphics[width=0.43\textwidth]{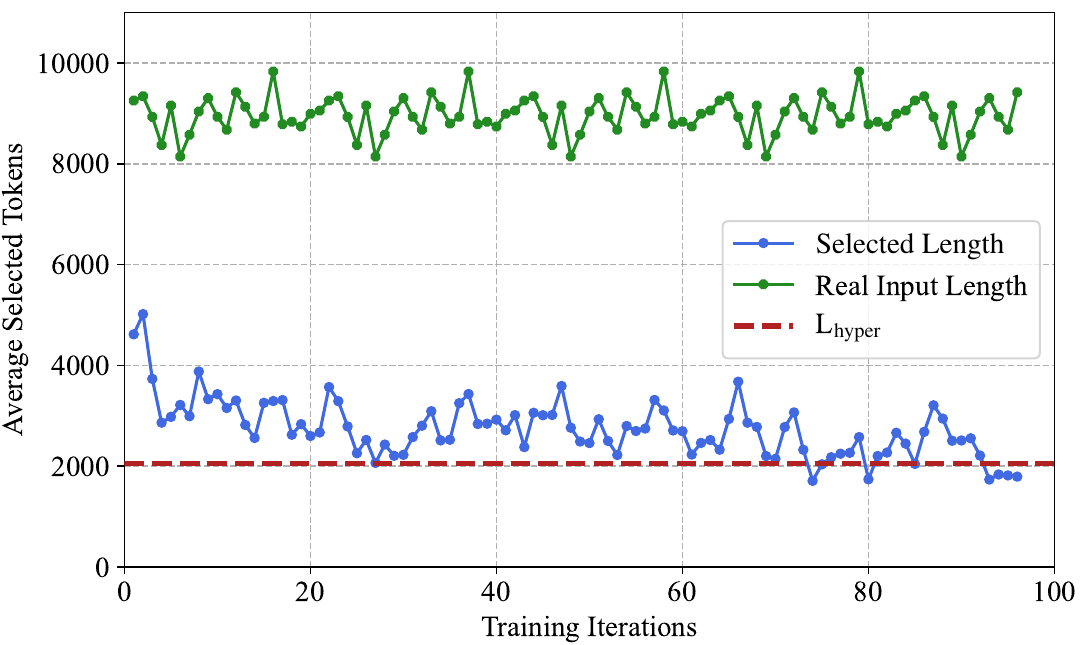}
    \includegraphics[width=0.46\textwidth]{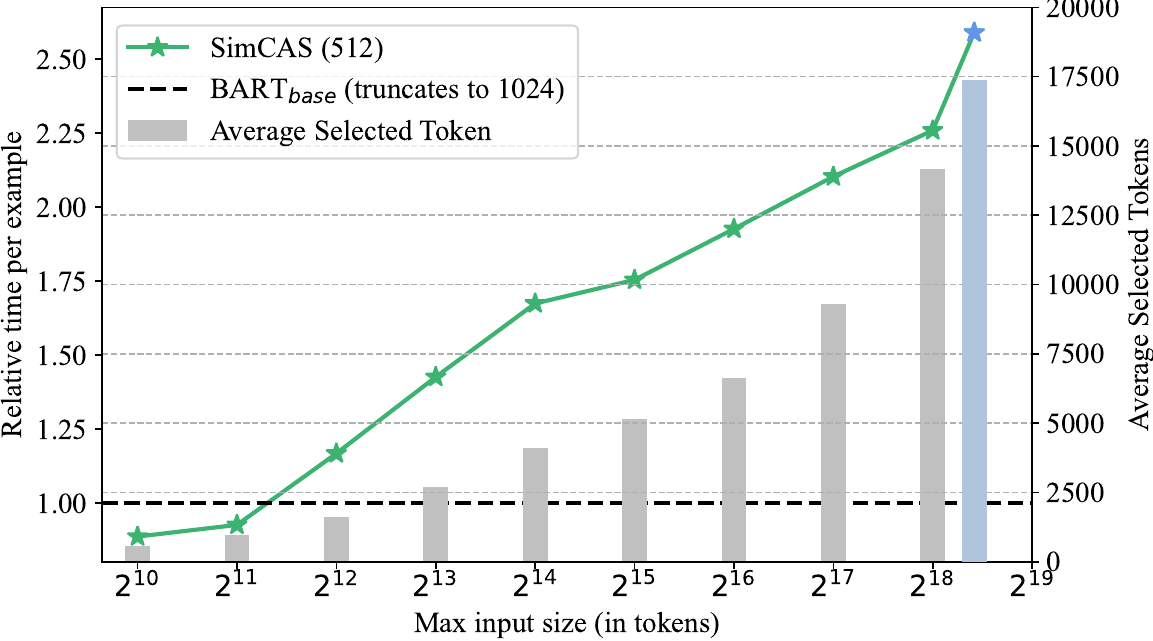}
    \caption{The left-hand-side plot shows the change of the actual number of tokens entered into the model and the number of tokens selected by our selector during training. The red dashed line represents the conditional boundary for the skipping reward. The right-hand-side plot shows the effect of increasing the number of input tokens in the inference phase on the time latency and the number of selected tokens. The area marked in blue on the right represents the limit of the number of tokens that the V100 can handle (350K tokens).}
    \label{fig:time_select}
\end{figure*}
Our implementation is based on \textit{PyTorch} \citep{NEURIPS2019_bdbca288} and \textit{Transformers} \citep{wolf-etal-2020-transformers} libraries. We train our model by using 8 NVIDIA V100 32G GPUs.

During the training phase, the maximum input lengths for BART$_\text{large}$ and BART$_\text{base}$ are set to 8192 and 16384, respectively, unless otherwise specified. To ensure efficient training, we update the parameters of the original backbone and the selector alternately. The reward estimation for each action is computed based on decoder cross-attention and the output feedback of the generative model. This estimation process is detached from the computation graph and does not participate in backpropagation.

At the inference stage, compared to the original generation process, our framework only adds a chunk-wise selection procedure between the encoder and the decoder, which takes very little time. At the decoding stage, the target sequence is generated with beam search in an auto-regressive manner~\citep{wiseman-rush-2016-sequence}.

\subsection{Evaluations}
Like most previous works, for abstractive summarization tasks, we measure the quality of generated summaries using the popular metric ROUGE \citep{lin-2004-rouge}. On the test set of arXiv, PubMed, GovReport, SummScreen, Multi-News, and WCEP, we report full-length F1-based ROUGE-1, ROUGE-2, and ROUGE-L scores computed with the standard ROUGE Perl package. Furthermore, we also use a popular model-based semantic metric BERTScore\footnote{In order to make a fair comparison with the previous work, we also use checkpoint ``facebook/bart-large-mnli'' for BERTScore.} \citep{Zhang*2020BERTScore:} to demonstrate the superiority of our approaches comprehensively. As shown in Table~\ref{tab:pearson}, ROUGE-1 and ROUGE-L are strongly correlated (Pearson correlation score of 0.913). In particular, the relatively low Pearson correlation coefficient ($<$ 0.8) between ROUGE-2 and BERTScore indicates a significant difference in preferences.
\par
As for the reading comprehension task, we use the F1 and exact match (EM) metrics defined in SCROLLS \citep{shaham-etal-2022-scrolls} to evaluate the model performance on the NarrativeQA dataset. Both metrics normalize the reference and system output strings via lowercasing, removing punctuation and stopwords, and normalizing whitespace.

\section{Discussion}

\begin{figure}
  \centering
  \includegraphics[width=0.45\textwidth]{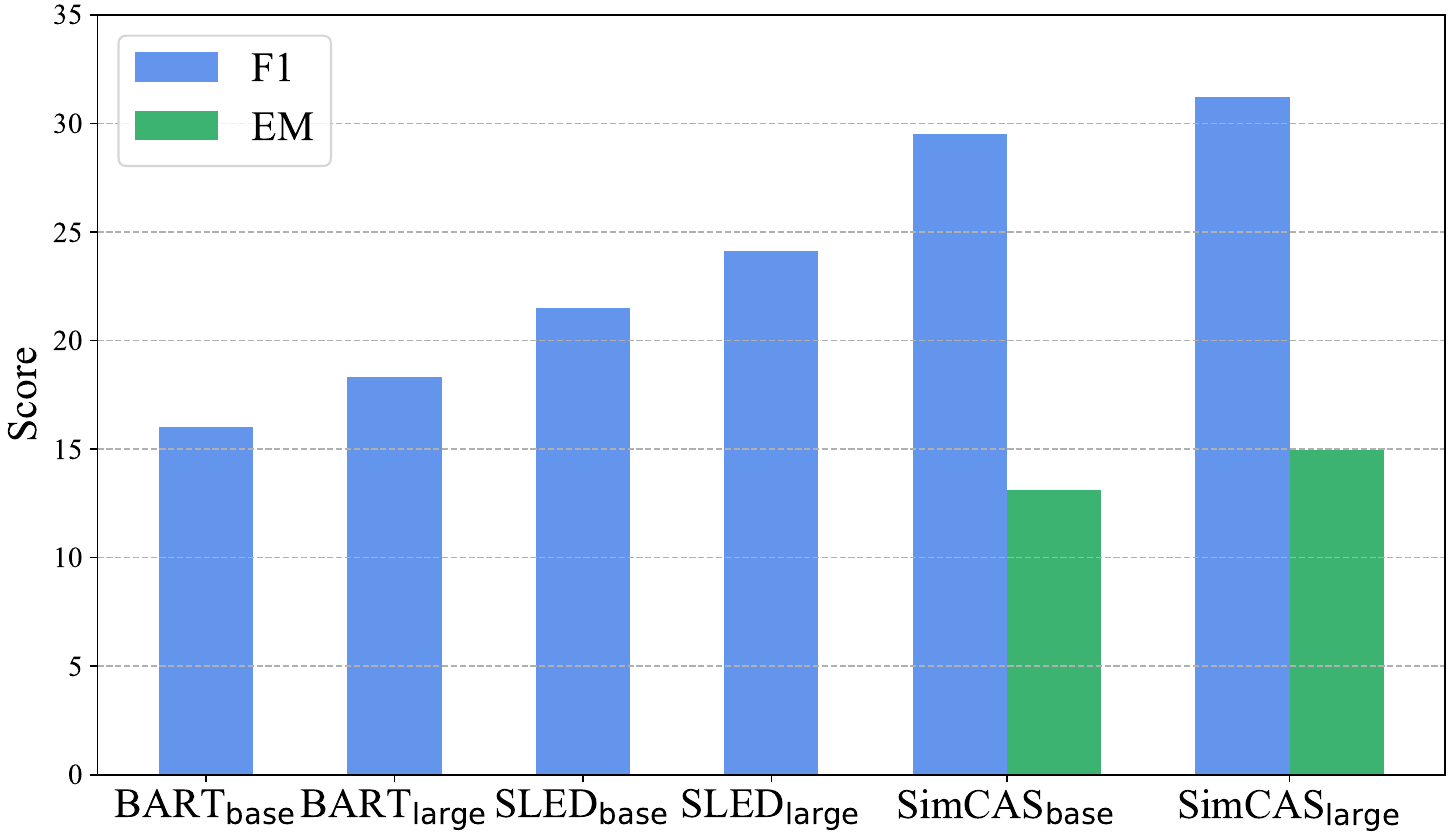}
  \vspace{-1mm}
  \caption{System performance comparison on NarrativeQA test set.}
  \label{fig:nrtv}
  \vspace{-2mm}
\end{figure}

\paragraph{Main Result Anaylses}
Table~\ref{tab:long} and \ref{tab:long-1} report results over four long-document test sets. We note that casting the backbone BART$_\text{base}$ into our SimCAS can significantly improve the model performance, and increasing the size of the model can further improve performance. Our approach outperforms baseline on several metrics and achieves new state-of-the-art performance on the PubMed test set. In Table~\ref{tab:multi-document}, the results of multi-document summarization tasks have a similar trend. Apart from PRIMERA, which customizes a pre-training objective for multi-document summarization, our method significantly outperforms previous results.
See Figure~\ref{fig:nrtv}, we observe a substantial improvement in the performance of our model on the NarrativeQA test set compared to previous works. Given the fact that this dataset has an extremely long average input length ($\geq$100K) and short average output length ($\leq$10), we attribute this significant performance enhancement to our method’s proficiency in filtering out an immense amount of task-irrelevant information. This, in turn, enables the decoding blocks to fuse information more efficiently.

\begin{figure*}[t]
    \centering
    \includegraphics[width=0.99\textwidth]{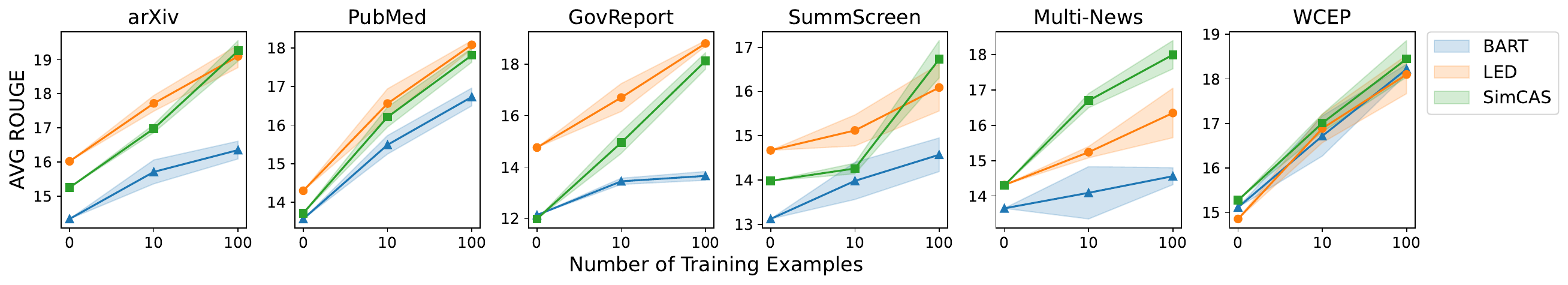}
    \vspace{-2mm}
    \caption{The AVG ROUGE scores (R-1, R-2, and R-L) of the pre-trained models  (BART$_\text{base}$, LED$_\text{base}$, SimCAS$_\text{base}$) with 0, 10, and 100 training examples with variance. All results are obtained by the average of 5 random runs with different seeds.}
    \label{fig:few_shot}
    \vspace{-3mm}
\end{figure*}

\paragraph{Ablation Study}
SimCAS is composed of three components: \texttt{Chunk}, \texttt{Align}, and \texttt{Select}. To investigate the contributions of each component, we independently removed each one. Notably, after removing \texttt{Chunk}, the maximum input length was restricted to 1024, causing \texttt{Align} to fail. As demonstrated in Table~\ref{tab:ablation}, the performance experienced a significant decline when either \texttt{Chunk} or \texttt{Select} was removed, underscoring their effectiveness. The \texttt{Align} component improved performance on all datasets, with the exception of the Multi-News dataset. We hypothesize that more sophisticated alignment methods could potentially enhance performance further.

\paragraph{Effectiveness of Conditional Skipping Reward} 
To prevent our selector from choosing an excessive number of token embeddings, we implement a conditional reward mechanism for skipping actions, which imposes a soft constraint on the number of selected token embeddings. Figure~\ref{fig:time_select} (left) demonstrates the effectiveness of this reward mechanism. As training progresses, the number of selected embeddings gradually converges to a threshold of 2048. This indicates that our selection process is effective in identifying essential token embeddings. Additionally, as the input sequence length increases, the ratio of selected token embeddings to the total length decreases. This is due to the length penalty in equation~\ref{eq:select_skip_reward}, which discourages the selection of an excessive number of token embeddings.

\paragraph{Increasing Maximum Input Length}
We explore the performance changes of our model on a single 32G V100 GPU as the input sequence length increases during inference. Initially, we assess the efficiency of our framework in processing long sequences using BART$_\text{base}$ as the backbone. We depict the fluctuations in the ratio of the average time required by SimCAS to the time taken by BART (1024) in Figure~\ref{fig:time_select} (right). Specifically, a line graph represents the relative time per sample, while a bar graph displays the number of token embeddings selected by SimCAS. When the input length is short ($<2^{14}$), the time cost exhibits near-linear growth. We attribute this phenomenon to our method's significant reduction in computational complexity and the redundant computation resource for short sequences. However, with a dramatic increase in input length ($>2^{17}$), the time cost rises sharply. As indicated in the blue area on the far right of the graph, our method can handle input sequences of up to 350k in length on a V100. Moreover, even as the number of input tokens increases significantly, the selected hidden states remain at a reasonable size, demonstrating our selector's capability to retrieve high-contribution information.

\paragraph{Low-Resource Evaluation} In real-world applications, the quantity of training samples available for downstream tasks is frequently quite constrained. Consequently, the performance of the model under low-resource conditions is of paramount importance. In our setup, we randomly sample a few (10 and 100) training examples from specific datasets to adapt these PLMs such as BART$_\text{base}$ (1024), LED$_\text{base}$ (4096), and SimCAS$_\text{base}$ (4096),  for corresponding data distributions. The results in Figure~\ref{fig:few_shot} indicate that our model demonstrates superior sample efficiency in low-resource scenarios compared to previous robust baselines.

\vspace{-.5mm}
\section{Conclusion and Future Work}
\vspace{-.5mm}
In this paper, we introduced a simple method for long-text processing via chunking, aligning, and selecting, called SimCAS. We divided the input long sequence into chunks, and encoded them with sequential batch alignment to capture the inter-chunk semantics. To select the important token representations in the encoded output of the encoder, we introduced a reinforcement learning-based token selector with the PPO method. 
We leverage the transformer as an environment and design a reward scheme for the corresponding actions based on the output logits and decoder cross-attention feedback to optimize the hidden token selector. Substantial experiment results and analyses demonstrate the satisfying effectiveness of SimCAS. Besides, our method does not depend on any particular tasks or models, which have good generalization ability for various application scenarios.

For future work,
our method can be naturally adopted into non-language long-sequence processing tasks, such as molecular structure analysis. 
Also, SimCAS has the potential to enhance the long text pre-training to transformers.

\vspace{-0.5mm}
\section{Limitations}
\vspace{-0.5mm}
Even though our framework significantly reduces the overhead of processing long sequences, due to memory constraints, the maximum input length on a single V100 32G GPU during training is 16k, and there is an extremely large consumption of GPU memory even if the batch size is small, which might suppress the potential of our method. In addition, we use small-scale PLMs with good performance as the backbone for downstream tasks. The effectiveness of our framework on large language models needs to be investigated further.

\section{Ethical Considerations}
\vspace{-.5mm}
Similar to existing methods, there is no guarantee that the generated content is factually consistent and free from hallucination \cite{maynez-etal-2020-faithfulness,kang-hashimoto-2020-improved}. Therefore caution is imperative when applying our method to scenarios with high demand of generation accuracy.
\bibliography{custom}
\clearpage
\appendix

\section{Dataset Statistics} \label{appx:statistics}
\begin{figure*}[htbp]
    \centering
    \includegraphics[width=0.98\textwidth]{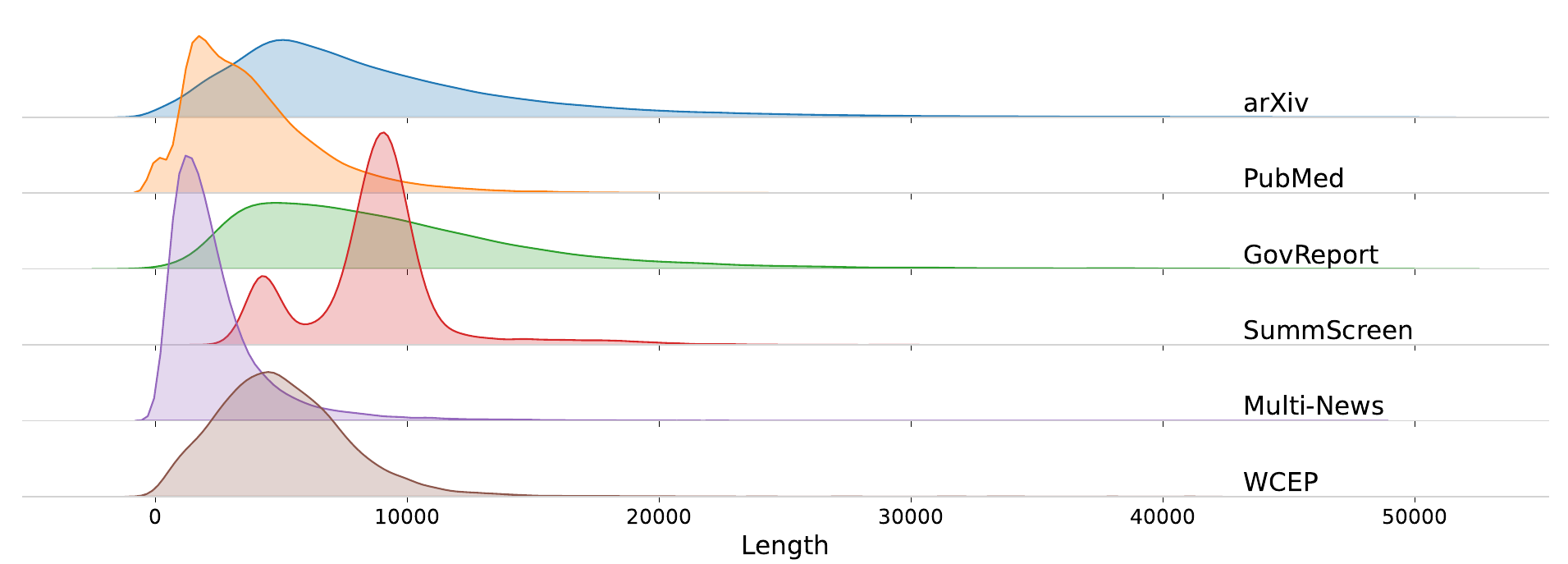}
    \caption{Input text length distributions on the six summarization datasets }
    \label{fig:kde}
\end{figure*}

\begin{table}[htbp] \small
    \centering
    \setlength{\tabcolsep}{1.2mm}
    \begin{tabular}{ c c c c c }
        \toprule
        Dataset & Train & Valid & Test & Avg. Input Tokens \\
        \midrule
        arXiv & 203.0K & 6.4K & 6.4K & 6.0K \\
        PubMed & 119.9K & 6.6K & 6.7K & 3.0K \\
        GovReport & 17.5K & 1.0K & 1.0K & 9.6K \\
        SummScreen & 22.6K & 2.1K & 2.1K & 6.6K \\
        Multi-News & 44.9K & 5.6K & 5.6K & 1.8K \\
        WCEP & 8.1K & 1.0K & 1.0K & 3.9K \\
        NarrativeQA & 32.7K & 3.5K & 10.6K & 121.7K \\
        \bottomrule
        \end{tabular}
        \caption{Statistics of used datasets.}
        \label{tab:datasets}
\end{table}



\begin{table*}[tbp]
    \centering
    \setlength{\tabcolsep}{3.0mm}
    \begin{tabular}{ l | c | c | c | c }
        \toprule
        \textbf{Dataset} & \textbf{Model} & \textbf{Batch Size} & \textbf{Max Epoch} & \textbf{Warmup Steps} \\
        \midrule
        arXiv & BART$_\text{base}$, BART$_\text{large}$ & 1 & 10 & 6400,10000 \\
        \midrule
        PubMed & BART$_\text{base}$, BART$_\text{large}$ & 1 & 10 & 6400, 10000 \\
        \midrule
        GovReport & BART$_\text{base}$, BART$_\text{large}$ & 1 & 20 & 6400, 10000\\
        \midrule
        SummScreen & BART$_\text{base}$, BART$_\text{large}$ & 1 & 20 & 2500, 6400 \\
        \midrule
        Multi-News & BART$_\text{base}$, BART$_\text{large}$ & 1 & 20 & 6400, 10000 \\
        \midrule
        WCEP & BART$_\text{base}$, BART$_\text{large}$ & 1 & 100 & 1600, 2500 \\
        \midrule
        NarrativeQA & BART$_\text{base}$, BART$_\text{large}$ & 1 & 10 & 6400, 10000\\
        \bottomrule
        \end{tabular}
        \caption{Hyper-parameter grid for downstream task fine-tuning. We use Adam optimizer ($\beta_1 = 0.9$, $\beta_2 = 0.999$, $\epsilon = 1e-6$) for all datasets.}
        \label{tab:hyper-1}
\end{table*}

\begin{table*}[tbp]
    \centering
    \setlength{\tabcolsep}{3.0mm}
    \begin{tabular}{ l | l}
        \toprule
        \textbf{Dataset} & \textbf{Generation Parameters}  \\
        \midrule
        arXiv & beam: 4, max\_len: 300, min\_len: 50, no\_repeat\_ngrams: 3, length\_penalty: 5.0\\
        \midrule
        PubMed & beam: 4, max\_len: 400, min\_len: 40, no\_repeat\_ngrams: 3, length\_penalty: 4.0\\
        \midrule
        GovReport & beam: 4, max\_len: 740, min\_len: 50, no\_repeat\_ngrams: 3, length\_penalty: 4.0\\
        \midrule
        SummScreen & beam: 4, max\_len: 750, min\_len: 50, no\_repeat\_ngrams: 3, length\_penalty: 4.0\\
        \midrule
        Multi-News & beam: 4, max\_len: 400, min\_len: 150, no\_repeat\_ngrams: 3, length\_penalty: 2.0\\
        \midrule
        WCEP & beam: 4, max\_len: 40, min\_len: 15, no\_repeat\_ngrams: 3, length\_penalty: 2.0\\
        \midrule
        NarrativeQA & beam: 4, max\_len: 20,  no\_repeat\_ngrams: 3, length\_penalty: 1.0\\
        \bottomrule
        \end{tabular}
        \caption{Hyper-parameter settings during inference for each dataset.}
        \label{tab:hyper-2}
\end{table*}

\section{Hyper-parameters \& Packages}

Table~\ref{tab:hyper-1} and Table~\ref{tab:hyper-2} delineate the hyper-parameter settings at the training and inference phases of the experiment, respectively. For evaluation metrics, we used the following packages:
\begin{itemize}[leftmargin=0.4cm]
    \item For the uniform data processing, before the evaluation, all the reference texts and model-generated texts are converted to the lowercase and tokenized using the PTB tokenizer: \url{https://nlp.stanford.edu/nlp/javadoc/javanlp/edu/stanford/nlp/process/PTBTokenizer.html}.
    \item For ROUGE metrics, we used the public rouge-score Perl package provided by the authors: \url{https://github.com/summanlp/evaluation/tree/master/ROUGE-RELEASE-1.5.5}.
    \item For BERTScore \cite{Zhang*2020BERTScore:}, we used the public bert-score package shared by the authors: \url{https://github.com/Tiiiger/bert_score}.
\end{itemize}

\section{Introduction for Baselines} \label{appx:baseline}

We use the competitive baselines that demonstrate the downstream task results for comparison. Among them, \textbf{BART} \citep{lewis-etal-2020-bart} is a standard full-attention PLM for sequence generation. Compared with BART, \textbf{PEGASUS} \citep{pmlr-v119-zhang20ae} has a tailored pre-training objective for abstractive text summarization. \textbf{LED} (Longformer Encoder-Decoder) \citep{beltagy2020longformer} uses the sparse attention-based encoder and full-attention decoder. Before pre-training, its parameters are initialized from BART. \textbf{BIGBIRD} \citep{zaheer2020big}, for an encoder-decoder setup, also introduces their specified sparse attention mechanism only at the encoder side.  \textbf{PRIMERA} based on LED introduces a task-specific per-training objective for multi-document summarization. \textbf{SLED} \citep{10.1162/tacl_a_00547} processes long sequences via short-context PLMs. The origin long sequence is partitioned into overlapping chunks. \textbf{HEPOS} \citep{huang-etal-2021-efficient} proposes head-wise positional strides to effectively pinpoint salient information from the source. \textbf{Memorizing Transformers} \citep{wu2022memorizing} employs a trainable attention gate to moderate between the standard cross-attention and attention over retrieved keys from a datastore. \textbf{Unlimiformer} \citep{bertsch2023unlimiformer} uses the k-nearest-neighbor KNN index to choose full-length input to reduce computation overhead. \textbf{HiMAP} \citep{fabbri-etal-2019-multi} expands the existing pointer-generator network model into a hierarchical network. \textbf{GraphSum} \cite{li-etal-2020-leveraging-graph} employs well-known graph representations of documents to effectively process multiple input documents for abstractive summarization. \textbf{BART-Long-Graph} \citep{pasunuru-etal-2021-efficiently} is fine-tuned based on LED and additionally injects discourse graph. \textbf{LED+RELAX} \cite{parnell2022multidocument} introduces a RELAX gradient estimator with multi-document coverage reward. \textbf{BERTREG} \citep{gholipour-ghalandari-etal-2020-large} is a regression-based sentence ranking system with BERT embedding, which is used as an extractive summarization method. \textbf{Submodular+Abs} \citep{gholipour-ghalandari-etal-2020-large} consists of a submodular-based extractive summarizer and a bottom-up abstractive summarizer. \textbf{DynE} \citep{hokamp2020dyne} ensembles multiple-document for abstractive summarization by single document summarization models.

\section{Perplexity on Test Sets}

To more comprehensively demonstrate the effectiveness of our approach, we additionally use perplexity as an evaluation metric for comparative experiments. The experimental results in Table \ref{tab:ppl} indicate that our system (SimCAS) significantly enhances model performance compared to the baselines (BART).

\begin{table*}[tbp]
    \centering
    \setlength{\tabcolsep}{1.0mm}
    \begin{tabular}{c c c c c c c c}
        \toprule
         \textbf{Method} & \textbf{arXiv} & \textbf{PubMed} & \textbf{GovReport} & \textbf{SummScreen} & \textbf{Multi-News} & \textbf{WCEP} & \textbf{NarrativeQA} \\
        \midrule
        BART & 31.82 & 34.81 & 42.10 & 48.91 & 49.40 & 50.91 & 23.34 \\
        SimCAS & 6.89 & 5.58 & 6.23 & 10.80 & 8.25 & 8.85 & 4.71 \\
        \bottomrule
        \end{tabular}
        \caption{Perplexity on all seven test sets including arXiv, PubMed, GovReport, SummScreen, Multi-News, WCEP, and NarrativeQA.}
        \label{tab:ppl}
\end{table*}

\section{Time Latency at Inference Stage}

At the inference stage, our framework additionally introduces a selection process compared with the standard transformer. Therefore, We study the proportion of our selector in the average inference time cost per sample. The results in Table \ref{tab:latency} indicate that our selector only adds a small amount of time cost while significantly improving model performance, demonstrating the effectiveness of our approach.

\section{Potential to select representation}

\begin{figure}[tbp]
    \centering
    \includegraphics[width=0.46\textwidth]{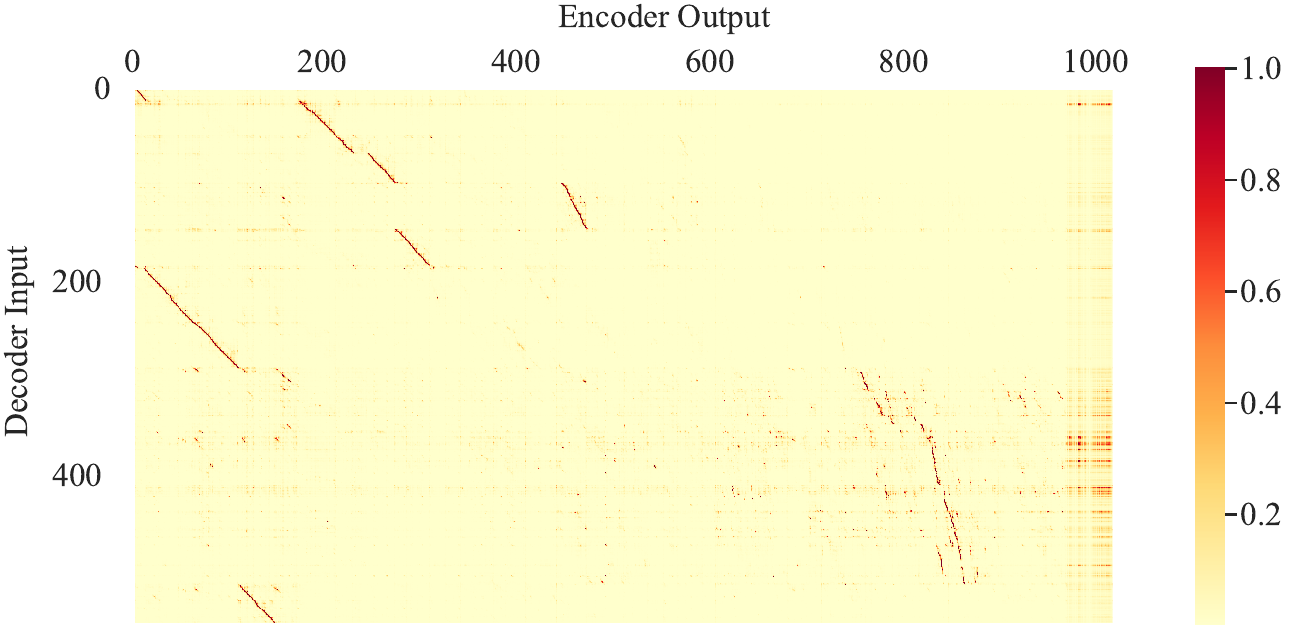}
    \caption{Visualization of the average cross-attention of all attention heads at all layers in the decoder (excluding start position). This example is generated by the BART$_\text{base}$+SimCAS trained on GovReport.}
    \label{fig:attention}
\end{figure}

Because in the decoder cross-attention module, the encoded output from the encoder is used as ``Key'' and ``Value'' to participate in the calculation, its attention score can lead to the contribution of the corresponding token representation from encoded output to the current token decision at the decoding step. Figure\ref{fig:attention} demonstrates how the cross-attention scores change during the decoding of the reference output with one example in GovReport. We can observe that 1) most of the token decisions in the decoding phase focus only on a small set of encoded representations; 2) For each token decision, the contributions of different encoded token representations vary greatly. These phenomena suggest that there is still the feasibility of further filtering out low-contribution encoded representations.

\section{Compatible with short-input tasks}
\begin{table*}[tbp]
    \centering
    \setlength{\tabcolsep}{1.0mm}
    \begin{tabular}{c c c c c c c c}
        \toprule
         \textbf{Method} & \textbf{arXiv} & \textbf{PubMed} & \textbf{GovReport} & \textbf{SummScreen} & \textbf{Multi-News} & \textbf{WCEP} & \textbf{NarrativeQA} \\
        \midrule
        Selection (ms) & 157.5 & 70.4 & 293.9 & 103.4 & 74.2 & 9.8 & 112.4 \\
        All (ms) & 9661.5 & 13799.1 & 104953.7 & 30424.0 & 17664.2 & 1356.2 & 689.6 \\
        Ratio (\%) & 1.63 & 0.51 & 0.28 & 0.34 & 0.42 & 0.72 & 16.30 \\
        \bottomrule
        \end{tabular}
        \caption{The average time overhead of the selection process (Select) and the whole process (All), and their ratio (Ratio) for a single sample.}
        \label{tab:latency}
\end{table*}

\begin{figure}[tbp]
    \centering
    \includegraphics[width=0.46\textwidth]{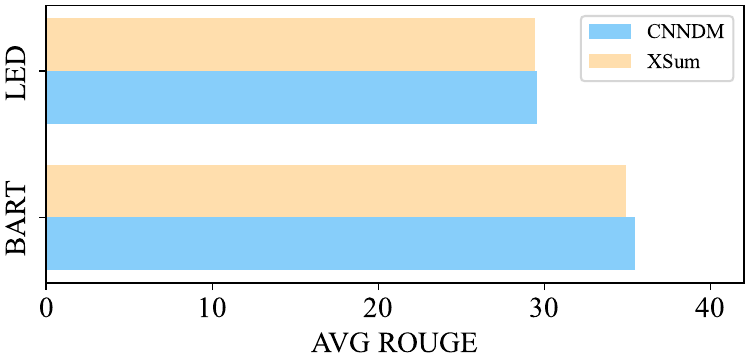}
    \caption{Comparison of the performance of full-attention PLM BART and sparse-attention PLM LED on short-text summarization datasets CNNDM \citep{nallapati-etal-2016-abstractive} and XSum \citep{narayan-etal-2018-dont}. AVG ROUGE denotes the average of ROUGE-1/2/L F1 scores.}
    \label{fig:long_short}
\end{figure}

While the sparse-attention transformers have been proven effective on a wide range of long-sequence datasets, as shown in Figure~\ref{fig:long_short}, these methods tend to underperform traditional full-attention transformers on the more common short-sequence tasks. However, in real scenarios, short-sequence inputs and long-sequence inputs are often mixed together, and the former occupies the vast majority. This limits the application scope of sparse-attention transformer architecture. In contrast, applying our framework SimCAS to existing full-attention transformers has strong flexibility. Specifically, given the full-attention model under SimCAS, if the input sequence exceeds the maximum length of a single chunk, we will perform chunking and selecting, otherwise, we can naturally switch to a standard short-text encoding form by skipping chunking procedures.

\section{Efficacy of SBA}
\begin{figure}[tbp]
    \centering
    \includegraphics[width=0.46\textwidth]{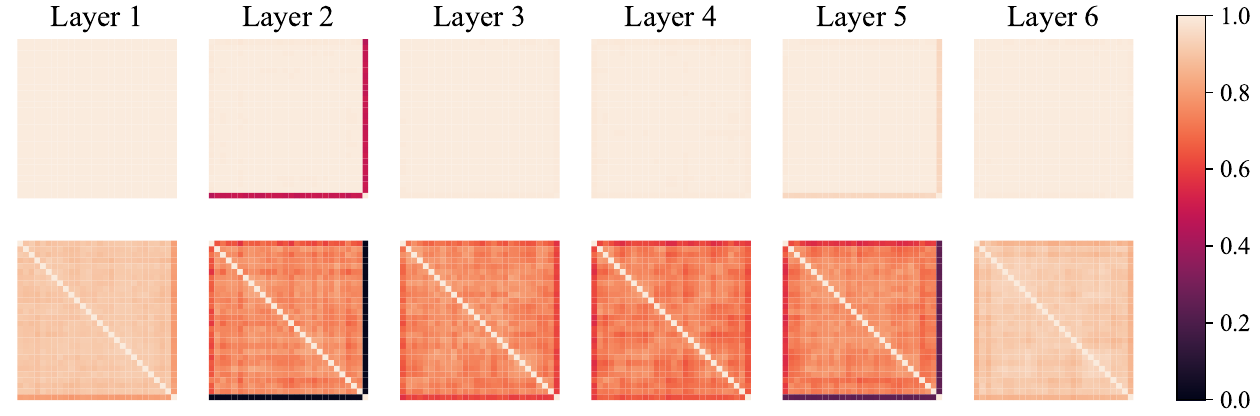}
    \caption{Evolution of similarities between start hidden states of each chunk after each encoding layer during one forward propagation. Only cosine similarity is used in the top half of the figure. The lower part takes into account both cosine similarity and Euclidean distance.}
    \label{fig:role}
\end{figure}

As our framework introduces SBA to align inter-chunk information during the encoding steps, we explore the change of hidden state in the forward propagation process under the influence of this component. Figure~\ref{fig:role} demonstrates the visualized similarity matrix between each chunk's start hidden state after the encoding block. It can be seen from an array of cosine similarity matrices in the top half that, through the alternation of SBA and encoding blocks, the cosine similarities between the start hidden states from different chunks are all close to 1, which indicates that their directions in high-dimensional space are almost the same.

Considering that cosine similarity can only reflect the closeness between vector directions, we design a similarity calculation method to add the measure of Euclidean distance. Given a pair of vectors $\vv_1, \vv_2 \in \mathbb{R}^{d}$, the custom similarity \textbf{Sim} between them is formulated as follows:

\begin{equation*}
    \textbf{Sim} = \frac{(\vv_1, \vv_2)}{\lVert\vv_1\rVert * \lVert\vv_2\rVert * (1 + \lVert\vv_1 - \vv_2\rVert)},
\end{equation*}\noindent
by which, we scale the cosine similarity to observe its change. As can be seen from the lower part of Figure~\ref{fig:role}, after each encoder layer, the start hidden states will vary in scale due to different contexts. 

\begin{algorithm}[tbp]
\caption{Token embedding selection with PPO}\label{alg:ppo}
\begin{algorithmic}[1]
\STATE Initialize agent and hyper-parameters
\STATE Initialize advantages, rewards, and returns
\STATE \textbf{Compute Rewards and Advantages}
\STATE Initialize next\_value $\leftarrow 0$, last\_gae\_lam $\leftarrow 0$
\STATE valid\_reward\_num $\leftarrow$ size of select\_rewards
\FOR{$t=0$ to num\_steps-1}
    \STATE Compute action sum for each step and update rewards
\ENDFOR
\STATE cur\_num $\leftarrow 0$
\FOR{$t=0$ to num\_steps-1}
    \STATE Update rewards and actions based on select\_rewards
\ENDFOR
\FOR{$t =$ num\_steps-1 to 0}
    \STATE Compute delta, advantages, and returns
\ENDFOR
\STATE Reverse advantages and returns

\STATE \textbf{Policy and Value Function Update}
\FOR{$epoch = 0$ to update\_epochs-1}
    \STATE Shuffle batch indices
    \FOR{$batch\_st = 0$ to seq\_len by mini\_batch}
        \STATE Compute policy loss, value loss, and entropy loss
        \STATE Perform gradient ascent
    \ENDFOR
    \IF{approx\_kl $>$ threshold}
        \STATE Break
    \ENDIF
\ENDFOR
\end{algorithmic}
\end{algorithm}

\section{More Experimental Details} \label{appx:settings}
Since we aim to minimize dataset-specific modeling, we unify the processing of the used long-document summarization, multi-document summarization, and machine reading comprehension datasets. In the NarrativeQA dataset for machine reading comprehension, we concatenate together the question and reference content, each of which prepends a prefix to indicate the type of text. For a fair comparison, we uniformly use the full-attention PLM BART\footnote{The checkpoints of BART are ``facebook/bart-base'' and ``facebook/bart-large'' containing around 139 M and 406 M parameters respectively, whose maximum encoding length is 1024.} on the seven public datasets above. Built on the BART, our framework introduces an additional parameterized selector to focus on more task-specific token representations. The selector follows the actor-critic style \citep{NIPS1999_6449f44a} and contains around 8M parameters. There are two Adam optimizers with $\beta_{1} = 0.9$, $\beta_{2} = 0.999$ for BART and selector respectively.
\par
Additionally, to handle longer sequences during the training phase, we set the chunk size to 512 instead of 1024 in all experiments (considering the start token and end token). To maintain chunk-wise alignment, we pad the chunks to uniform the size of each chunk. During the forward propagation, the embedding layer embeds position representations for each chunk independently.\footnote{Considering the context of each chunk is different, although our design may lead to repeated position representation, we argue that after the encoding stage, the combination of the same token and position information would still produce various representations \citep{kazemnejad2023impact}}

\paragraph{Training Details}
In this paper, for the training of BART$_\text{base}$ on various datasets we uniformly employ the Adam optimizer with the following dynamic learning rate:

\begin{small}
\begin{equation*}
    lr = \gamma \min (\textit{step}^{-0.5}, \textit{step} \times \textit{warmup}^{-1.5}),
\end{equation*}
\end{small}\noindent
where $\gamma$ affects the maximum learning rate and \textit{warmup} indicates the warmup steps, \textit{step} is the number of updating steps, and $lr$ is the learning rate. In addition, there is another separate Adam optimizer for our reinforcement learning-based parameterized selector, in which the learning rate is fixed to $1 \times 10^{-4}$. The optimization of the selector and transformer is performed alternately, with one model trained while the other model is fixed.

\paragraph{Inference Details}
In practice, We also investigate how the model performs when the maximum effective input length is increased during inference. In Table~\ref{tab:length}, we can observe that although the maximum input length is set to 16384 during the training phase due to the memory limitation, increasing this maximum length during inference still improves the model performance.
\begin{table}[tbp] \small
    \centering
    \setlength{\tabcolsep}{1mm}
    \begin{tabular}{c c c c c}
        \toprule
         \textbf{Maximum Input Length} & \textbf{R-1} & \textbf{R-2} & \textbf{R-L} & \textbf{BS} \\
        \midrule
        4096 & 44.96 & 22.60 & 37.17 & 69.74 \\
        8192 & 45.41 & \textbf{22.79} & 37.40 & 70.51 \\
        16384$^{\star}$ & 45.22 & 22.48 & 37.34 & 70.46 \\
        32768 & 45.45 & 22.56 & 37.45 & 70.52 \\
        $+\infty$ & \textbf{45.68} & 22.80 & \textbf{37.71} & \textbf{70.59} \\
        \bottomrule
        \end{tabular}
        \caption{Performance of BART$_\text{base}$-SimCAS on WCEP with different maximum input lengths during inference. R-1/2/L is the ROUGE-1/2/L F1 score. BS refers to the model-based metric BERTScore. $\star$: the maximum input length in the training phase. $+\infty$: the input sequence to the model is complete. The best results are bolded.}
        \label{tab:length}
\end{table}

\paragraph{The Details of Selector}
Our token embedding selection process based on PPO is shown in Algorithm~\ref{alg:ppo}. In our setting, the selector consists of an actor component and a critic component with PPO optimization, both of which are the feed-forward network, except for the final layers. In order to enable the selector to choose more diverse token representations instead of becoming homogeneous during the chunk-wise selection process, the state space consists of the current token representation and the selector's hidden state that is affected by previous actions. At the beginning of the selection procedure, the initial selector hidden state is the average of the start hidden state of all chunk representations. At each time step, we input the current selector hidden state and a chunk of token representations. Then the actor of the selector outputs a probability distribution over action space, and the critic of the selector outputs a single estimated scalar value for each token based on the hidden state selector and the corresponding token representation. For the state transition after executing current actions, we update the hidden state of the selector using the selected token representation from the current chunk. Note that in order to avoid the extreme case where the selector skips all tokens, in each chunk decision, if all actions are ``\textit{skipping}'', they are all switched to ``\textit{selecting}'', and the corresponding action probability and value are re-obtained. Additionally, to increase the training stability, the advantage in the PPO algorithm is approximated using the Generalized Advantage Estimation (GAE) \citep{DBLP:journals/corr/SchulmanMLJA15}.
\section{Case Study on GovReport Dataset}

\begin{figure*}[tbp]
    \centering
    \includegraphics[width=0.99\textwidth]{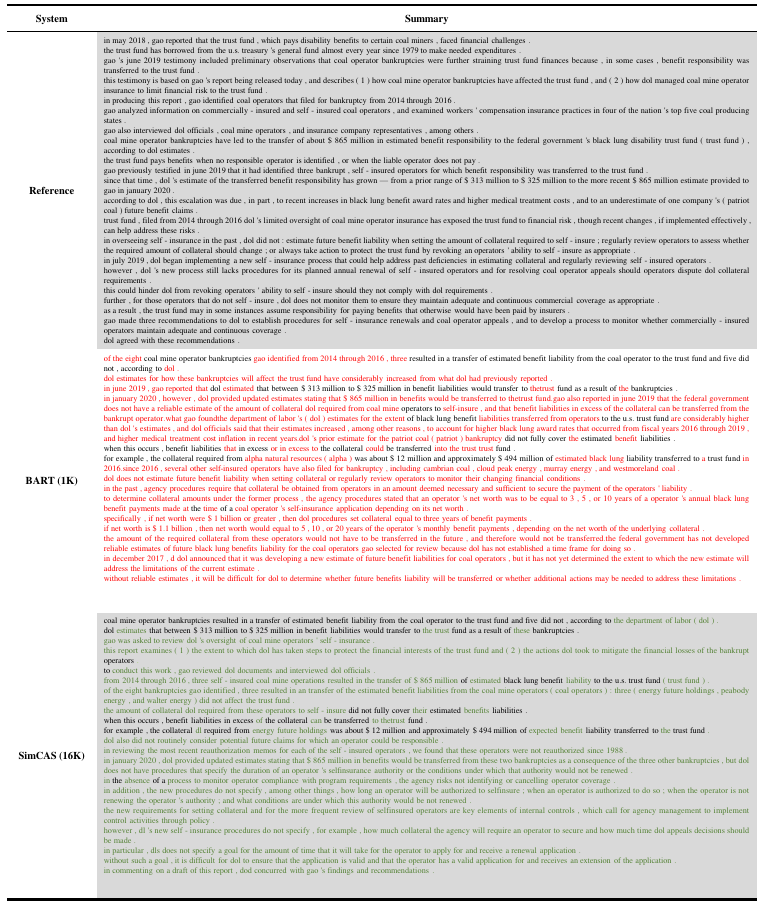}
    \caption{Example summary generated by BART and SimCAS trained on GovReport dataset. The maximum input length of standard BART and SimCAS is 1024 and 16384 respectively. The sentence in green is included in the SimCAS summary, while the one in red is discarded.}
    \label{fig:case}
\end{figure*}

In addition to using regular automatic evaluation metrics to measure the effect of model generation, we also present some actual output to support the results. Figure~\ref{fig:case} displays several examples of summaries generated by the fine-tuned base model BART$_\text{large}$ and our BART$_\text{base}$-SimCAS. We can observe that the system output of our model has fewer grammatical errors and higher coherence compared with the base model. Furthermore, since our model is able to perceive longer sequences, our output is more informative and better aligned with the reference text.

\end{document}